\documentclass[sigconf,nonacm]{acmart}

\AtBeginDocument{%
  }

\renewcommand\footnotetextcopyrightpermission[1]{}
\begin{document}

\title{FIDAC: An Easy-to-use Pipeline to Extract and Interpret
  Interpersonal Distance From Video}

\author{Keshav Rastogi}
\affiliation{%
  \institution{University of California, Berkeley}
  \city{Berkeley}
  \state{CA}
  \country{United States}
}

\author{Eugy Han}
\affiliation{%
  \institution{Department of Media Production, Management, and Technology, University of Florida}
  \city{Gainesville}
  \state{FL}
  \country{United States}
}

\author{Jeremy N. Bailenson}
\affiliation{%
  \institution{Department of Communication, Stanford University}
  \city{Stanford}
  \state{CA}
  \country{United States}
}

\renewcommand{\shortauthors}{Rastogi et al.}

\begin{abstract}
The distance between persons reveals significant information about their perception of each other. However, such information is not easily
extractable and interpretable from video input. We developed an open-sourced library, Facial Interpersonal Distance Analysis and Coding
(FIDAC) that transforms facial detection results into actionable data about location and interpersonal distance. This tool merges data from
multiple open-source facial detection models, strategically compensating for gaps in any individual model. In addition, we include methods for more accurate tracking, such as a pipeline for human coding of the selection of faces and a benchmarking tool to reduce depth distortion.
For next steps, we plan on building upon FIDAC by evaluating its effectiveness at measuring interpersonal distance at various depths and
orientations while further integrating features of proxemic analysis such as synchrony into its software.
\end{abstract}


\maketitle




\section{Introduction}

Interpersonal distance (IPD), the space people maintain between each other, reveals significant information about their behavior and perception of others. Past research has shown that, at closer distances, individuals tend to disclose more information, collaborate more effectively, and experience greater comfort and attentiveness
\cite{JourardFriedman1970, Quintard2018, Birtchnell1993}. Such an effect simultaneously exists in reverse: when individuals feel more trusting and comfortable with someone else, they are likely to move closer \cite{Rosenberger2020, Bryan2012}. However, at extremely close distances, interactions may appear intrusive and a violation of personal space, prompting people to move away and re-establish their personal boundaries \cite{Lloyd2009, Hayduk1983}. Together, these
observations can be summarized by Edward Hall's \cite{Hall1966} four zones of IPD: intimate (0--46 cm), personal (47--120 cm), social (121--370 cm), and public (380--760 cm).

IPD has been a crucial variable of interest for scholars seeking to explore the nuances of human behavior and social engagement. The most
prominent method to measure it is the stop-distance technique \cite{Hayduk1978, Rosenberger2020}, where the distance between a
subject and an experimenter is slowly decreased until the subject feels uncomfortable \cite{Perry2013, Bailenson2003}. For example, a subject may start to approach a digital agent in virtual reality and indicate to a researcher when they do not want to move any closer \cite{Kroczek2020}. While effective and relatively simple to set up, the stop-distance technique can only capture a subject's initial response, which may pose limitations as IPD may fluctuate throughout an interaction.

To tackle this problem, many researchers have opted to either digitally or manually track a subject's motion throughout their interaction.
Recent digital innovations include virtual and mixed reality, where headsets can capture the motion data of a subject's head throughout the
experiment \cite{Bailenson2003, Han2023}. Furthermore, advances in motion-tracking technology have made it possible for researchers to
quantify specific movements without the use of markers or wearable sensors \cite{Ito2022, Das2023}. Such technologies, however, face
numerous limitations: they can be expensive, difficult to operate, or hard to acquire \cite{Romero2017}.

Other forms of tracking software may face further barriers. For instance, when a subject's hair may cover their face or their body position may be oriented away from the sensor, these segments of data may be omitted \cite{Baumann2011}. As such, incorporating human coding to account for potential errors may enhance the reliability of the
data. However, this process tends to either be tedious, costing hours of video review, or inaccessible for researchers that may
lack the necessary programming experience.

Hence, there is a need for a digital program that can easily yet effectively measure a subject's position while enabling a researcher to
fill in gaps in cases of errors. With a sequence of a subject's position, researchers may then analyze associated social nonverbal
cues, such as interpersonal distance or synchrony. This goal motivated the development of Facial Interpersonal Distance Analysis and Coding
(or FIDAC), an easy-to-use program that allows researchers to investigate interpersonal distance.

\section{Methodology}

\begin{figure}[h]
  \centering
  \includegraphics[width=\linewidth]{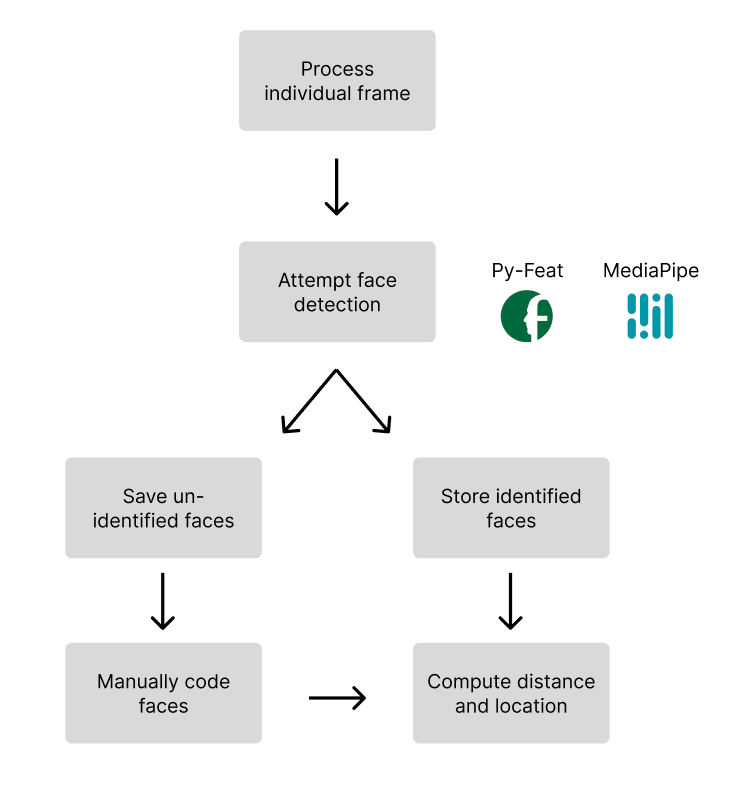}
  \caption{FIDAC's three-step pipeline for analyzing interpersonal distance: frame extraction, facial detection (model-generated and human-coded), and data merging}
  \label{fig:pipeline}
\end{figure}

FIDAC utilizes a three-step process to analyze interpersonal distance (see Figure~\ref{fig:pipeline}). First, video data is analyzed and each frame is extracted and processed through OpenCV. Second, facial detection is attempted. If a face is present, the data (facial location and facial confidence score) is stored in a CSV file. If a face is not present, a separate manual encoding file is created, allowing the researcher to revisit the same data and human-code such instances later. Finally, the manual, human-coded data is merged with the model-generated data, offering the researcher a chance to evaluate the data and conduct further analysis.

\subsection{Video Setup}

FIDAC requires video input from a smartphone or digital camera. Researchers are encouraged to position these sources to best capture
the desirable range of facial motion. While FIDAC's algorithms were trained to identify faces from varying angles and orientations \cite{Namba2021, Kollias2018}, we note that it is less efficient if obstacles like long hair or masks block a face at difficult angles. Although FIDAC can standardize position through the benchmarking tool, since it primarily computes relative position, it is important to maintain a stable position of video input which can be achieved through methods like a tripod.

\subsection{Analysis + Facial Detection}

FIDAC begins by extracting every frame from an input video using OpenCV. However, to boost efficiency and speed, only a certain number
of frames are extracted per second to analyze differences. For instance, one may choose to only evaluate one or sixty frames per second.

Then, a facial detection model identifies the location of faces present in the frame. FIDAC comes pre-loaded with Py-Feat \cite{Cheong2021} and
Google's MediaPipe \cite{Lugaresi2019}, but researchers can add their own facial detection model. A researcher may also choose to use only
one of the models, or a combination of both to ensure that discrepancies in one are corrected by another. This provides an additional benefit to running each source of input through a model individually.

\begin{figure}[h]
  \centering
  \includegraphics[width=\linewidth]{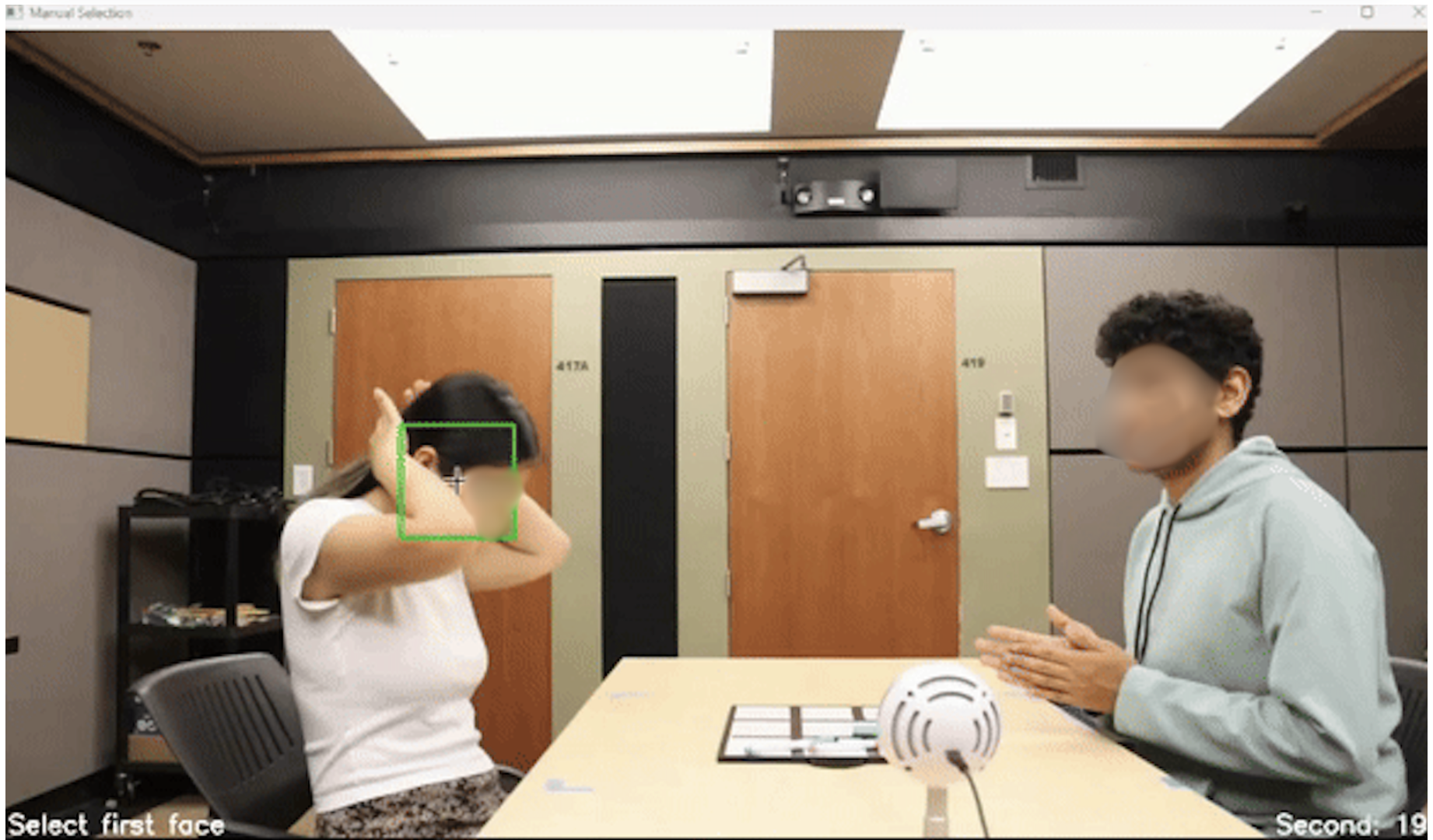}
  \caption{A detected face represented as a bounding box}
  \label{fig:boundingbox}
\end{figure}

Each face is represented by a rectangle, often referred to as a bounding box, that most facial detection models output (see Figure~\ref{fig:boundingbox}). Each detected face is also assigned a score or confidence value. Enabling a high threshold of confidence is
recommended when setting up the model to ensure that only faces above the threshold are detected.

To mitigate the potential for false positives, FIDAC allows for the setting of limits for the sizes of faces detected. This way, even if a
false positive is detected, if its dimensions do not meet the size limits, it is not counted as a face.

\subsection{Benchmarking}

FIDAC is designed for 2D recordings, such as the output from a smartphone or digital camera. While convenient and easy to operate, these devices cannot determine depth from an image. Especially when computing interpersonal distance, accounting for depth is important to both accurately determine a subject's position and interpret the effects of such a distance, such as Hall's zones of proximity
\cite{Hall1966}.

\begin{figure}[h]
  \centering
  \includegraphics[width=\linewidth]{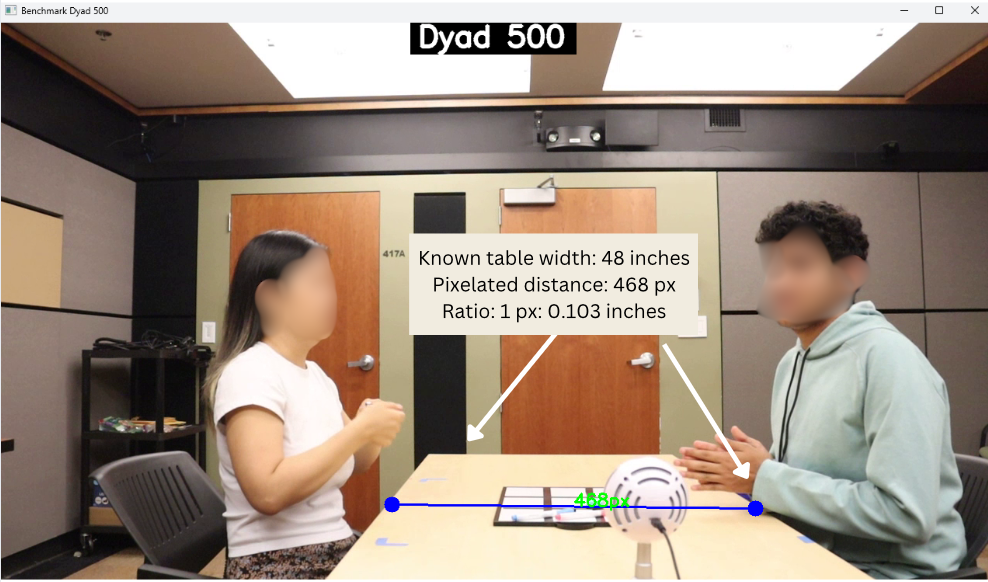}
  \caption{Benchmarking tool used to convert pixelated distances into real-world units by referencing two points on the same plane as the participants for which the true distance is known}
  \label{fig:benchmarking}
\end{figure}

FIDAC loops through an input video and enables a researcher to pause and select two points on the same plane as the participants for which
the distance is known. For instance, if participants are seated opposite each other on a table, a researcher could select both sides of
the table and use the known distance as a ratio against the pixelated distance FIDAC calculates (see Figure~\ref{fig:benchmarking}).

\section{Conclusion and Future Directions}

Facial Interpersonal Distance Analysis and Coding (FIDAC) was designed to alleviate the lack of an accessible, easy-to-use system to analyze
dynamic human motion and interpersonal distance. Toward this goal, we designed a system that integrates numerous facial detection algorithms
to track a participant's motion from a 2D video input, such as from a smartphone or digital camera. Additionally, researchers can substitute
their own models, modifying settings to correct discrepancies in any individual model. For instances where a model fails, FIDAC includes a
fully built pipeline for researchers to human-code the remaining data. Lastly, to translate the pixelated distances from FIDAC's facial detection into unitary values, a benchmarking tool is provided.

After developing a foundational core of FIDAC, we intend on further developing and testing its capabilities. In particular, we envision
creating a dataset and evaluating FIDAC's performance at various depths and orientations against existing facial detection algorithms. We also plan on conducting a user study, working with scholars researching interpersonal communication to explore how they use FIDAC and its benefits. Based on these findings, we hope to evolve FIDAC into a complete pipeline, enabling scholars to upload a video and automatically receive insights about IPD or synchrony.

\subsection*{Acknowledgements}
This project was supported by the Sloan Foundation and the Data Sciences Institute at the University of Toronto with funding through grant number DSI-QRSY3R1P03.

\bibliographystyle{ACM-Reference-Format}
\bibliography{references}

@article{JourardFriedman1970,
  author  = {Jourard, Sidney M. and Friedman, Robert},
  title   = {Experimenter-Subject "Distance" and Self-Disclosure},
  journal = {Journal of Personality and Social Psychology},
  year    = {1970},
  volume  = {15},
  number  = {3},
  pages   = {278--282}
}

@article{Quintard2018,
  author  = {Quintard, Virginie and Jouffre, St{\'e}phane and Croizet, Jean-Claude and Bouquet, Cedric A.},
  title   = {The Influence of Passionate Love on Self-Other Discrimination during Joint Action},
  journal = {Psychological Research},
  year    = {2020},
  volume  = {84},
  number  = {1},
  pages   = {51--61},
  doi     = {10.1007/s00426-018-0981-z}
}

@book{Birtchnell1993,
  author    = {Birtchnell, John},
  title     = {How Humans Relate: A New Interpersonal Theory},
  year      = {1993},
  publisher = {Praeger},
  address   = {Westport, CT}
}

@article{Rosenberger2020,
  author  = {Rosenberger, Lisa A. and Naef, Michael and Eisenegger, Christoph and Lamm, Claus},
  title   = {Interpersonal Distance Adjustments after Interactions with a Generous and Selfish Trustee during a Repeated Trust Game},
  journal = {Journal of Experimental Social Psychology},
  year    = {2020},
  volume  = {90},
  pages   = {104001}
}

@article{Bryan2012,
  author  = {Bryan, Ronnie and Perona, Pietro and Adolphs, Ralph},
  title   = {Perspective Distortion from Interpersonal Distance Is an Implicit Visual Cue for Social Judgments of Faces},
  journal = {PLOS ONE},
  year    = {2012},
  volume  = {7},
  number  = {9},
  pages   = {e45301}
}

@article{Lloyd2009,
  author  = {Lloyd, Donna M.},
  title   = {The Space between Us: A Neurophilosophical Framework for the Investigation of Human Interpersonal Space},
  journal = {Neuroscience \& Biobehavioral Reviews},
  year    = {2009},
  volume  = {33},
  number  = {3},
  pages   = {297--304}
}

@article{Hayduk1978,
  author  = {Hayduk, Leslie A.},
  title   = {Personal Space: An Evaluative and Orienting Overview},
  journal = {Psychological Bulletin},
  year    = {1978},
  volume  = {85},
  number  = {1},
  pages   = {117--134}
}

@article{Hayduk1983,
  author  = {Hayduk, Leslie A.},
  title   = {Personal Space: Where We Now Stand},
  journal = {Psychological Bulletin},
  year    = {1983},
  volume  = {94},
  number  = {2},
  pages   = {293--335}
}

@book{Hall1966,
  author    = {Hall, Edward T.},
  title     = {The Hidden Dimension},
  year      = {1966},
  publisher = {Doubleday},
  address   = {Garden City, NY}
}

@article{Perry2013,
  author  = {Perry, Anat and Rubinsten, Orly and Peled, Leehe and Shamay-Tsoory, Simone G.},
  title   = {Don't Stand So Close to Me: A Behavioral and ERP Study of Preferred Interpersonal Distance},
  journal = {NeuroImage},
  year    = {2013},
  volume  = {83},
  pages   = {761--769}
}

@article{Bailenson2003,
  author  = {Bailenson, Jeremy N. and Blascovich, Jim and Beall, Andrew C. and Loomis, Jack M.},
  title   = {Interpersonal Distance in Immersive Virtual Environments},
  journal = {Personality and Social Psychology Bulletin},
  year    = {2003},
  volume  = {29},
  number  = {7},
  pages   = {819--833}
}

@article{Kroczek2020,
  author  = {Kroczek, Leon O. H. and Pfaller, Michael and Lange, Bastian and M{\"u}ller, Mathias and M{\"u}hlberger, Andreas},
  title   = {Interpersonal Distance during Real-Time Social Interaction: Insights from Subjective Experience, Behavior, and Physiology},
  journal = {Frontiers in Psychiatry},
  year    = {2020},
  volume  = {11},
  pages   = {561}
}

@article{Han2023,
  author  = {Han, Eugy and Miller, Mark R. and DeVeaux, Cyan and Jun, Hanseul and Nowak, Kristine L. and Hancock, Jeffrey T. and Ram, Nilam and Bailenson, Jeremy N.},
  title   = {People, Places, and Time: A Large-Scale, Longitudinal Study of Transformed Avatars and Environmental Context in Group Interaction in the Metaverse},
  journal = {Journal of Computer-Mediated Communication},
  year    = {2023},
  volume  = {28},
  number  = {2},
  pages   = {zmac031}
}

@article{Ito2022,
  author  = {Ito, Naoaki and Sigur{\dh}sson, Haraldur B. and Seymore, Kayla D. and Arhos, Elanna K. and Buchanan, Thomas S. and Snyder-Mackler, Lynn and Silbernagel, Karin Gr\"avare},
  title   = {Markerless Motion Capture: What Clinician-Scientists Need to Know Right Now},
  journal = {JSAMS Plus},
  year    = {2022},
  volume  = {1},
  pages   = {100001}
}

@article{Das2023,
  author  = {Das, Kishor and de Paula Oliveira, Thiago and Newell, John},
  title   = {Comparison of Markerless and Marker-Based Motion Capture Systems Using 95\% Functional Limits of Agreement in a Linear Mixed-Effects Modelling Framework},
  journal = {Scientific Reports},
  year    = {2023},
  volume  = {13},
  pages   = {22880}
}

@article{Romero2017,
  author  = {Romero, Veronica and Amaral, Joseph and Fitzpatrick, Paula and Schmidt, R. C. and Duncan, Ashley W. and Richardson, Michael J.},
  title   = {Can Low-Cost Motion-Tracking Systems Substitute a Polhemus System when Researching Social Motor Coordination in Children?},
  journal = {Behavior Research Methods},
  year    = {2017},
  volume  = {49},
  number  = {2},
  pages   = {588--601}
}

@inproceedings{Baumann2011,
  author    = {Baumann, Jonathan and Kr{\"u}ger, Bj{\"o}rn and Zinke, Alexander and Weber, Andreas},
  title     = {Data-Driven Completion of Motion Capture Data},
  booktitle = {Proceedings of the Eurographics/ACM SIGGRAPH Workshop on Posters and Demos},
  year      = {2011}
}

@article{Namba2021,
  author  = {Namba, Shushi and Sato, Wataru and Osumi, Masaki and Shimokawa, Koh},
  title   = {Assessing Automated Facial Action Unit Detection Systems for Analyzing Cross-Domain Facial Expression Databases},
  journal = {Sensors},
  year    = {2021},
  volume  = {21},
  number  = {12},
  pages   = {4222}
}

@misc{Kollias2018,
  author       = {Kollias, Dimitrios and Zafeiriou, Stefanos},
  title        = {Aff-Wild2: Extending the Aff-Wild Database for Affect Recognition},
  year         = {2018},
  howpublished = {arXiv:1811.07770}
}

@article{Cheong2021,
  author  = {Cheong, Jin Hyun and Xie, Tiankang and Byrne, Sophie and Chang, Luke J.},
  title   = {Py-Feat: Python Facial Expression Analysis Toolbox},
  journal = {arXiv preprint arXiv:2104.03509},
  year    = {2021}
}

@misc{Lugaresi2019,
  author       = {Lugaresi, Camillo and Tang, Jiuqiang and Nash, Hadon and McClanahan, Chris and Uboweja, Esha and Hays, Michael and Zhang, Fan and Chang, Chuo-Ling and Yong, Ming Guang and Lee, Juhyun and Chang, Wan-Teh and Hua, Wei and Georg, Manfred and Grundmann, Matthias},
  title        = {MediaPipe: A Framework for Building Perception Pipelines},
  year         = {2019},
  howpublished = {arXiv:1906.08172}
}

\end{document}